\documentclass[11pt,a4paper]{article}
\usepackage{times,latexsym}
\usepackage{url}
\usepackage[T1]{fontenc}
\usepackage[acceptedWithA]{tacl2018v2}
\usepackage{amsmath}
\usepackage{graphicx}
\usepackage{booktabs}
\usepackage{bbm}
\usepackage{tabularx,ragged2e}
\usepackage{multicol,multirow}
\usepackage[labelformat=simple]{subcaption}
\usepackage[ruled,vlined]{algorithm2e}

\usepackage{xspace,mfirstuc,tabulary}

\newif\iftaclinstructions
\taclinstructionstrue 
\iftaclinstructions
\renewcommand{\confidential}{}
\renewcommand{\anonsubtext}{Neeraj Varshney, Swaroop Mishra, Chitta Baral \\Arizona State University \\
    \texttt\{nvarshn2, srmishr1, chitta\}@asu.edu}
\newcommand{\instr}
\fi

\iftaclpubformat 

\else

\fi

\title{Interviewer-Candidate Role Play: Towards Developing Real-World NLP Systems}

\author{
        Neeraj Varshney, 
        Swaroop Mishra, 
        Chitta Baral 
\\Arizona State University \\
    \texttt\{nvarshn2, srmishr1, chitta\}@asu.edu
}

\date{}

\begin{document}
\maketitle
\begin{abstract}

Standard NLP tasks do not incorporate several common real-world scenarios such as seeking clarifications about the question, taking advantage of clues, abstaining in order to avoid incorrect answers, etc. 
This difference in task formulation hinders the adoption of NLP systems in real-world settings.
In this work, we take a step towards bridging this gap and present a multi-stage task that simulates a typical human-human questioner-responder interaction such as an interview.
Specifically, the system is provided with question simplifications, knowledge statements, examples, etc. at various stages to improve its prediction when it is not sufficiently confident.
We instantiate the proposed task in Natural Language Inference setting where a system is evaluated on both in-domain and out-of-domain (OOD) inputs.
We conduct comprehensive experiments and find that the multi-stage formulation of our task leads to OOD generalization performance improvement up to 2.29\% in Stage 1, 1.91\% in Stage 2, 54.88\% in Stage 3, and 72.02\% in Stage 4 over the standard unguided prediction.
However, our task leaves a significant challenge for NLP researchers to further improve OOD performance at each stage.
\footnote{Code and datasets for our task are available at  \href{https://github.com/nrjvarshney/interviewer-candidate-role-play}{https://github.com/nrjvarshney/interviewer-candidate-role-play}}

\end{abstract}

\section{Introduction}
\label{introduction}

Despite impressive progress made in Natural Language Processing (NLP), we are far from employing these systems reliably in real-world tasks. 
This can be partially attributed to the misalignment between formulations of real-world and standard NLP tasks.
Specifically, real-world tasks present several scenarios that are often not included in the standard task formulations such as 
(1) seeking clarifications about the question
(2) taking advantage of clues provided at inference time
(3) learning from a few examples similar to the given question
(4) abstaining in order to avoid incorrect predictions, etc.

\begin{figure}[t]
    \centering
    \includegraphics[width=\linewidth]{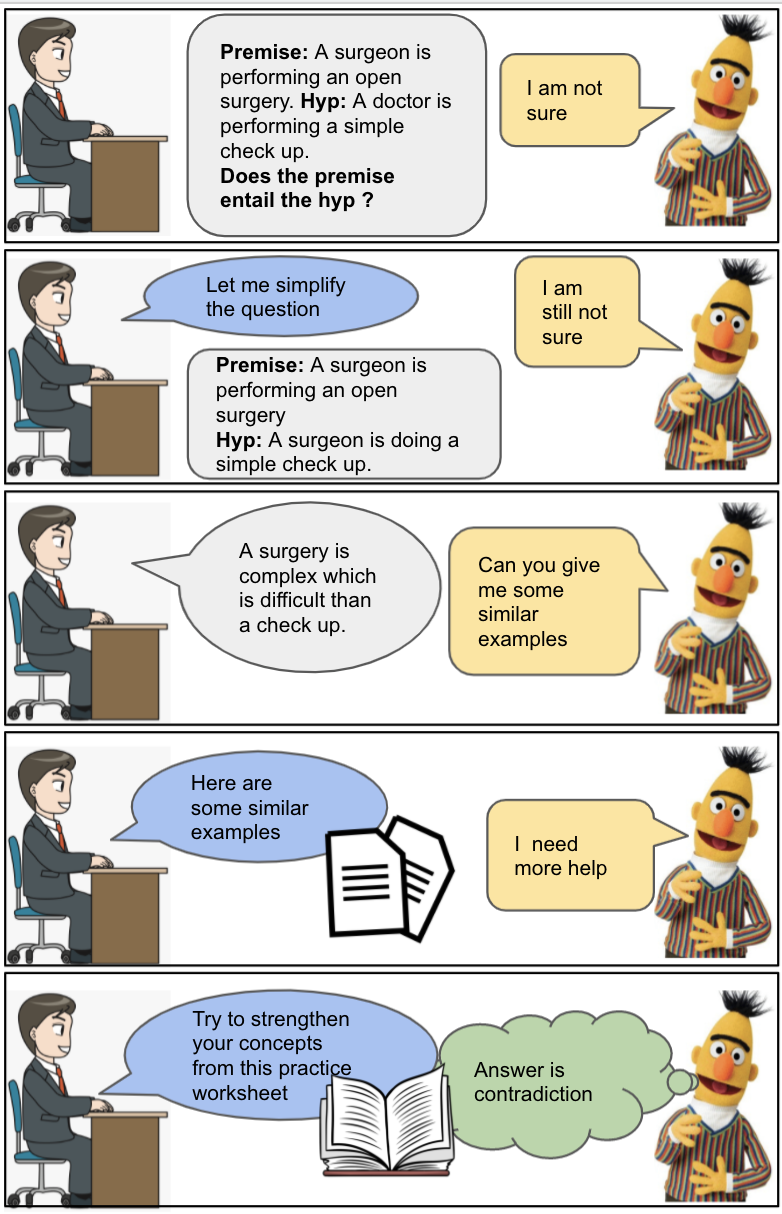}
    \caption{Illustration of the proposed task simulating an interviewer-candidate interaction for NLI.
    }
    \label{fig:intro_figure}
\end{figure}

In order to bridge this alignment gap, prior work in NLP has investigated few tasks that are closer to the real-world settings such as Selective Prediction \cite{kamath-etal-2020-selective,jones2020selective,varshney2020s}, Few-Shot Learning \cite{NEURIPS2020_1457c0d6, schick-schutze-2021-just, Ye2021CrossFitAF, Tam2021ImprovingAS}, Prompting \cite{shin-etal-2020-autoprompt,jiang-etal-2020-know,le-scao-rush-2021-many, Mishra2021NaturalIB}, etc.
Selective Prediction enables a system to maintain high accuracy by abstaining on instances where it is likely to be incorrect.
Few-Shot Learning challenges a system to learn from a limited number of training examples.
Prompts provide task/instance related guidance in order to improve model's predictions.
Though these works are a step in the right direction, they have several limitations. 
First, all these tasks give only a single opportunity to the system to either make a correct prediction or abstain. Whereas, in a typical human-human interaction, the questioner often gives hints, clarifications, examples, etc. in cases where the responder is not confident of their answer. 
Second, evaluation on these tasks is limited to specific aspects of system performance.
This motivates research into designing a realistic task that simulates a questioner-responder interaction and provides a unified evaluation of multiple aspects.

An interview is a prototypical example of questioner-responder interaction.
In an interview, when the candidate is not confident in their answer, the interviewer first tries to simplify the question in order to help them understand it better.
If the simplification doesn't help then they usually give some hints.
Next, they typically provide some similar examples as further assistance to improve their answer.
Finally, the interviewer may give a worksheet that has a number of similar unsolved questions and allow some more time for the candidate to strengthen their concepts and reattempt the question.

In this work, we present a multi-stage task that simulates the above-mentioned interviewer-candidate interaction as illustrated in Figure \ref{fig:intro_figure}.
Prior work has shown that model's confidence of prediction is often positively correlated with correctness similar to humans \cite{hendrycks17baseline,NIPS2017_9ef2ed4b} i.e a prediction is more likely to be correct if the confidence is high and more likely to be incorrect if the confidence is low.
Hence, we assist the system with input simplifications, clues, examples, etc. at various stages when it is not sufficiently confident in its prediction (Section \ref{proposed_task}). 
We organize our task into four sequential stages as shown in Figure \ref{fig:formalization}.
Initially, the model makes a prediction on the given test instance.
If the prediction confidence is below a certain threshold then it enters the first stage where a few semantic preserving simplifications of the test instance are provided.
The system is expected to utilize this help and better its prediction. 
If it enters Stage 2 then it is provided with some knowledge statements about the test instance.
Similarly, in the third stage, a few similar labeled examples are provided.
Finally, a number of unlabeled examples are given as further assistance in the fourth stage.
This task not only simulates a real-world scenario but also integrates a number of paradigms such as Selective Prediction, Prompting, Few-Shot Learning, and Unsupervised Learning.

We instantiate the proposed task in Natural Language Inference (NLI) setting (Section \ref{instantiation}) and evaluate on both in-domain and out-of-domain inputs.
We conduct comprehensive experiments on several NLI datasets and show that it improves OOD generalization performance up to 2.29\% in Stage 1, 1.91\% in Stage 2, 54.88\% in Stage 3, and 72.02\% in Stage 4 over the standard unguided prediction. 

In summary, our contributions are as follows:\\
(1) Addressing limitations of the standard NLP tasks, we propose a novel multi-stage task that is closer to the real-world setting and simulates an interviewer-candidate interaction.\\
(2) To the best of our knowledge, we are the first to study post-abstention scenarios where a model is assisted with guidance in various forms to answer the originally abstained questions. \\
(3) Our task improves OOD generalization performance up to 2.29\% in Stage 1, 1.91\% in Stage 2, 54.88\% in Stage 3, and 72.02\% in Stage 4 on the evaluation metric of the proposed task. There exists a noteworthy headroom for performance improvement on our task, which hopefully will motivate further work in this direction of developing NLP systems that align well with the real-world tasks. \\

\section{Related Work}
\subsection{Selective Prediction}
Selective Prediction task expects a system to answer when it likely to be correct and abstain otherwise.
There exists a large body of work on selective prediction in machine learning \cite{chow1957optimum,el2010foundations,geifman2017selective}.
Typically, the prediction confidence is used to decide when to answer and when to abstain.
In NLP, selective prediction has mostly been studied in connection with Calibration \cite{platt1999probabilistic} i.e aligning a model's output probability with the true probability of its predictions. 
\citet{desai-durrett-2020-calibration} study calibration of recently introduced pre-trained transformer models.
\citet{kamath-etal-2020-selective} train a calibrator leveraging softmax probabilities and instance-specific features such as input lengths for Question Answering (QA) models. 
\citet{varshney2020s} propose to transform calibration from classification to a regression problem incorporating difficulty scores of the instances.
\citet{zhang2021knowing} incorporate input example embedding from a pre-trained language model as additional features for the calibrator.
Unlike prior work, we also focus on post-abstention scenarios where a system is provided guidance in various forms to answer the originally abstained questions.

\subsection{Few-shot Learning}
Standard supervised learning approaches require a huge amount of labeled training data.
Inspired by human learning from just a few examples, few-shot learning presents a challenge of learning from a limited number of labeled training examples \cite{NEURIPS2020_1457c0d6, schick-schutze-2021-just, Ye2021CrossFitAF, Tam2021ImprovingAS}. Several works have shown that models can achieve comparable performance just by using a few representative samples \cite{wang2018dataset, nachum2018data, Mishra2020DoWN,sucholutsky2021less}.
Stage 3 in our task presents a few-shot learning challenge where a few labeled examples similar to the test instance are provided.

\subsection{Knowledge Addition}
Incorporating knowledge in models has been a long-standing research area in NLP.
Researchers leverage large knowledge banks such as COMET-ATOMIC \cite{hwang2020comet}, ConceptNet \cite{speer2017conceptnet}, etc. to solve commonsense reasoning tasks \cite{Mitra2019ExploringWT,chang-etal-2020-incorporating,shen-etal-2020-exploiting,Mishra2020TowardsQF} like CommonsenseQA \cite{talmor-etal-2019-commonsenseqa}, QUOREF \cite{dasigi-etal-2019-quoref}, etc.
Recently, prompting where additional task-related information is provided gained attention especially for regimes where only a small labeled dataset is available for training \cite{shin-etal-2020-autoprompt,schick-schutze-2021-exploiting,le-scao-rush-2021-many,Mishra2021NaturalIB}.
In our task, we provide instance-specific knowledge in Stage 2 when the system's confidence in its prediction is below a certain threshold.
\begin{figure}[t]
    \centering
    \includegraphics[width=\linewidth]{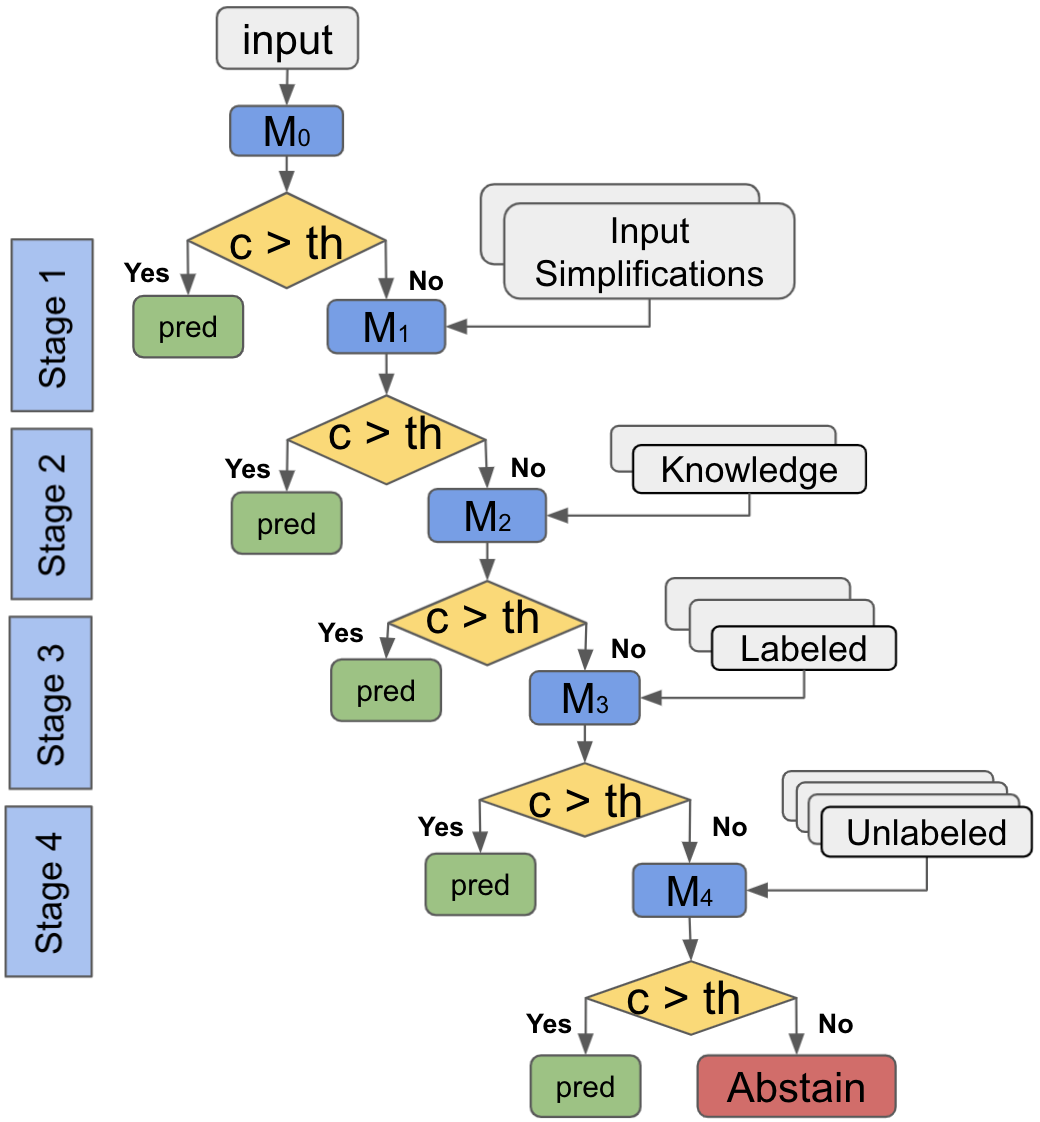}
    \caption{Illustration of various branches of the proposed task. Given an input, the model makes a prediction $pred$ with confidence $c$ sequentially in each stage leveraging the provided guidance until $c$ surpasses the threshold $th$. Note that $pred$ and $c$ vary with the stage but the threshold $th$ remains the same across stages.}
    \label{fig:formalization}
\end{figure}
\subsection{Unsupervised Learning}
Unsupervised Learning pertains to learning from unlabeled data \cite{lewis-etal-2019-unsupervised}.
This field is gaining interest as obtaining labeled data is both time consuming and expensive. In contrast, unlabeled data can be collected cheaply.
For downstream tasks, it has mostly been explored for Question Answering task ~\cite{chung-etal-2018-supervised,yang-etal-2017-semi, dhingra-etal-2018-simple,wang-jiang-2019-explicit,alberti-etal-2019-synthetic} where it is modeled as a data augmentation or a domain adaptation problem.
In this work, we provide unlabeled examples to the system in the final stage of our task.

\section{The Proposed Task}
\label{proposed_task}
In this section, we detail our proposed multi-stage task, its mathematical formulation, and evaluation metric.

\paragraph{Task Description:}
Prior work has shown that the model's confidence is often positively correlated with its correctness \cite{hendrycks17baseline,NIPS2017_9ef2ed4b} i.e its prediction is more likely to be correct if the confidence is high and more likely to be incorrect if the confidence is low.
Following this, we design our task in four sequential stages where a system goes from one stage to the next if it is not sufficiently confident in its prediction. Figure \ref{fig:formalization} illustrates the flow of the task. 
Each stage provides instance-specific guidance in various forms to assist the model in improving its prediction. 
If the prediction confidence in a stage exceeds a certain threshold then it attempts the test instance and skips the subsequent stages.

The stages are organized based on the steps that a typical interviewer follows in an interaction with a candidate (Section \ref{introduction}).
Initially, the model makes a prediction on the given input and the overall system outputs the prediction if the confidence is above a certain threshold and enters Stage 1 otherwise.
In Stage 1, it is provided with several semantic-preserving simplifications of the test instance.
The system is expected to leverage these input simplifications and improve its prediction on the given test instance.
Prior work has shown that even state-of-the-art models are sensitive to the input and simplifying the input can significantly boost model's performance \cite{jiang-etal-2020-know,Elazar2021MeasuringAI,anantha-etal-2021-open}.
In Stage 2, it is given some knowledge statements relevant to the test instance. 
A few similar labeled examples are provided in Stage 3.
In the final stage, it is further given a number of similar unlabeled examples. 
If the system fails to surpass the confidence threshold even after the final stage then it abstains from answering on that test instance in order to avoid incorrect prediction.

\paragraph{Mathematical Formulation:}
Algorithm \ref{algo:algo} shows the general structure of the proposed task.
\newcommand{\pluseq}{\mathrel{+}=}
\begin{algorithm}
\SetAlgoLined
\textbf{Given:} \\
$i$: Test Instance, \\ 
$th$: Confidence Threshold, \\
$M_0$: Trained Model, \\
$\mathcal{T}_s$: Stage-specific Guidance Function \\
\textbf{Initialization:} stage: $s \leftarrow  0$\\
\While {$s \leq 4 $}
{
    $M_s \leftarrow$ Update $M_{s-1}$ using $\mathcal{T}_s(i)$ if $s > 0$\\
    $pred_{i_s}, conf_{i_s} = M_s(i)$\\
    \uIf{$conf_{i_s} > th$}
    {
        \textbf{return} $pred_{i_s}$
    }
    $s += 1$
}
\textbf{return} ``Abstain''
 \caption{Task Structure}
 \label{algo:algo}
\end{algorithm}
The system continues to make the prediction on the given test instance leveraging the provided guidance until its confidence exceeds a certain threshold.
\citet{hendrycks17baseline} showed that $MaxProb$ (maximum softmax probability) is a simple yet strong estimate of prediction confidence.
Formally, $MaxProb$ estimates confidence on input $i$ as:
\begin{equation*}
    conf_{MaxProb} = \max_{y'\in Y(i)} Model(y'|i)
\end{equation*}
where $Y(i)$ denotes the possible output classes.

Calibration using a held-out dataset can further align the model's output probabilities \cite{lee2017training, kamath-etal-2020-selective} and give better confidence estimates.
The function $\mathcal{T}_s$ provides guidance to the system in Stage $s$ for an instance $i$ and is defined as:
\begin{equation*}
        \mathcal{T}_s(i) =
            \begin{cases}
              \emptyset, & \text{if $s = 0$} \\
              Simplified\ Inputs, & \text{if $s = 1$} \\
              Knowledge\ Stmts., & \text{if $s = 2$} \\
              Similar\ Labeled\ Ex., & \text{if $s = 3$} \\
              Similar\ Unlabeled\ Ex. & \text{if $s = 4$} \\
              
            \end{cases}
\end{equation*}


\paragraph{Evaluation Metric:}
\label{evaluation_metric}
In Selective Prediction, ``Coverage" is defined as the fraction of examples answered by the system while accuracy on covered examples is the fraction answered correctly. 
Furthermore, risk pertains to the error on the covered examples.
Selection of confidence threshold above which the system answers is application dependent i.e for tolerant applications like movie recommendation, a low threshold can be selected but for intolerant applications like medical diagnosis, a high threshold is selected to minimize risk. 
Hence, instead of evaluating a system at a particular threshold value, coverage and its associated risk is computed for every threshold value $th$ in order to estimate its overall performance.
As $th$ decreases, coverage will increase, but the risk will usually also increase. 
We plot risk versus coverage for all values of $th$ and calculate the area under this curve (AUC). AUC represents the overall performance of a method as it combines performance across all $th$ values. 
Lower AUC is preferred as it represents lower average risk across all thresholds.
We compute AUC for each stage as described below:

Let the model's initial prediction on instance $i$ be $pred_{i_0}$ with a confidence of $conf_{i_0}$ and prediction in stage $s = 1..4$ be $pred_{i_s}$ with a confidence of $conf_{i_s}$.
Note that the system gets to make a prediction in a stage only if the confidence in the previous stage was below the threshold $th$.
For every $th$ and stage $s$, we compute two values $c_{i_s}$ and $p_{i_s}$ as:
\begin{equation*}
        c_{i_s} =
            \begin{cases}
              c_{i_{s-1}}, & \text{if $conf_{i_{s-1}}$ > th} \\
              conf_{i_s} & \text{otherwise}
            \end{cases}
\end{equation*}
\begin{equation*}
        p_{i_s} =
            \begin{cases}
              p_{i_{s-1}}, & \text{if $conf_{i_{s-1}}$ > th} \\
              pred_{i_s} & \text{otherwise}
            \end{cases}
\end{equation*}
We use $c_{i_s}$ and $p_{i_s}$ to compute Coverage $C$ and Accuracy $A$ on covered examples in stage $s$ as:
\begin{equation*}
    C_s = \frac{\sum_{i=1}^n \mathbbm{1}(c_{i_s} \ge th)}{n} 
\end{equation*}
\begin{equation*}
    A_s =\frac{ \frac{1}{n}\sum_{i=1}^n (\mathbbm{1}(c_{i_s} \ge th)*v_{i_s})}{C_s} 
\end{equation*}

where, $\mathbbm{1}$ is the indicator function, $n$ is size of test dataset and parameter $v_{i_s}$ is $1$ when prediction $p_{i_s}$ is correct and 0 otherwise.  \\
We then plot risk-coverage curves and compute AUC to evaluate a system's performance.

\section{Task Instantiation}
\label{instantiation}
\begin{figure*}[t]
    \centering
    \includegraphics[width=0.8\textwidth]{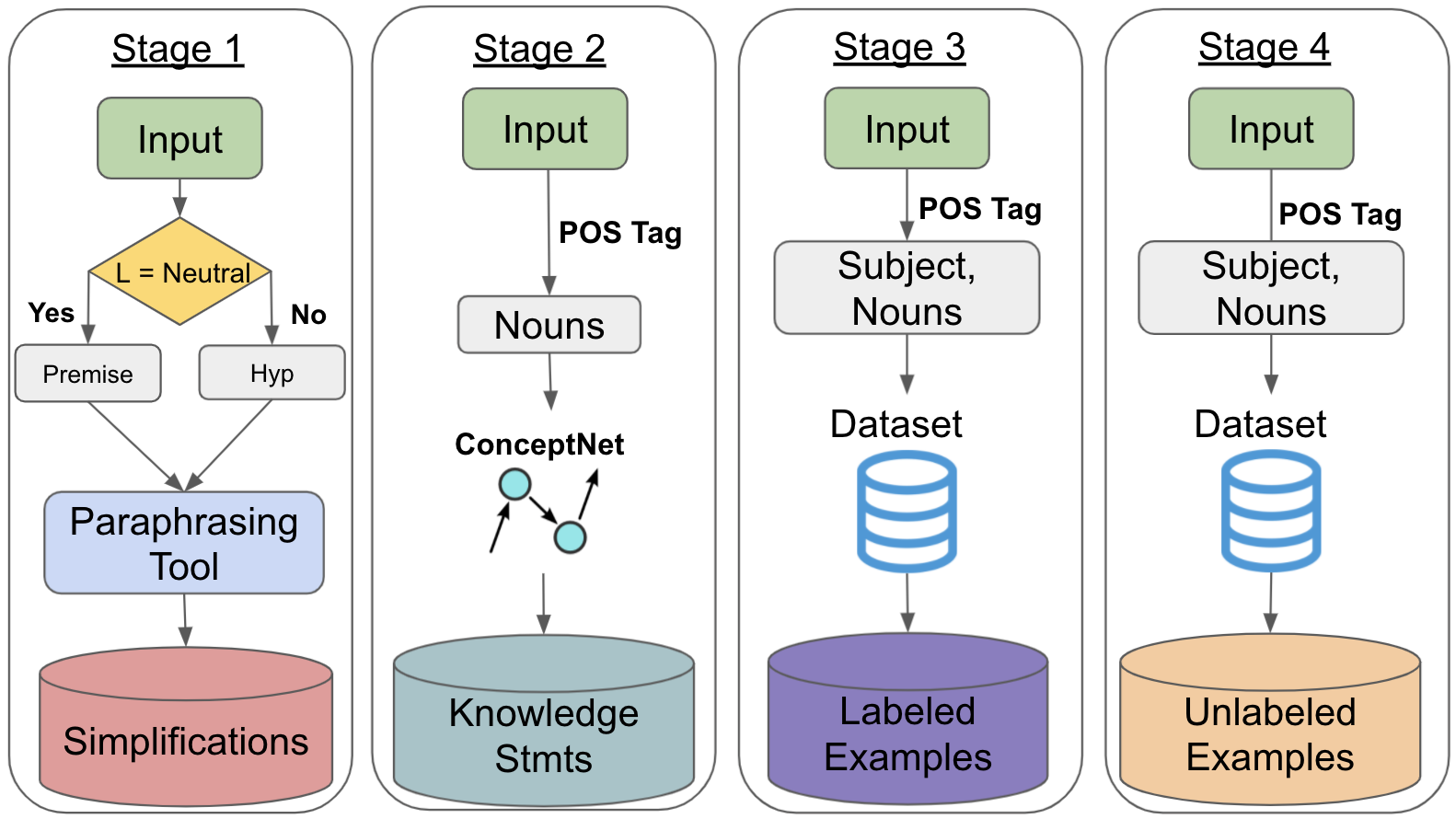}
    \caption{Steps involved at every stage during instantiation of the proposed task for NLI.}
    \label{fig:task_instantiation}
\end{figure*}

While our framework is general, we instantiate the proposed task in Natural Language Inference (NLI) that pertains to the task of identifying the relationship between a ``premise'' and a ``hypothesis'' sentence.
This relationship can be classified as either \textit{Entailment} (hypothesis must be true if the premise is true), \textit{Contradiction} (hypothesis can never be true if the premise is true), or \textit{Neutral} (hypothesis can be both true and false as the premise does not provide enough information to make a decision).
Table ~\ref{tab:nli_examples} shows examples of the NLI task.
We include the standard NLI datasets in our setup and consider SNLI \cite{bowman-etal-2015-large} as in-domain dataset with Multi-NLI \cite{williams-etal-2018-broad} and Dialogue NLI \cite{welleck-etal-2019-dialogue} as Out-of-Domain (OOD) datasets. 
We include the OOD datasets in our setup as the inputs often diverge from the model's training data in a real-world task.
Figure \ref{fig:task_instantiation} illustrates the method we followed to create various stages in our task instantiation.
We detail the four stages of the proposed task for NLI below:

\paragraph{Stage 1:}
The first stage requires the input to be simplified in case of abstention. 
We compile simplified versions of the premise-hypothesis tuples in an automated way using the paraphrasing tool introduced in \cite{zhang2020pegasus}.
We conduct a small user study in order to find the best strategy of employing this tool to compile semantic preserving variations of the original input.
We provide three sets of examples where we paraphrase only premise, only hypothesis, and both premise and hypothesis for the three labels (Entailment, Contradiction, and Neutral) separately.
We also provide the original PH tuple to the participants and ask them to annotate whether the transformation is label preserving.
We find that for Entailment and Contradiction PH tuples, paraphrasing the hypothesis is label preserving in most cases, whereas it is paraphrasing the premise for Neutral. 
Using these findings, we compile $10$ semantic preserving variations of the test instances and provide them to the system in case of abstention in the first stage.
Table ~\ref{tab:stage1_examples} shows examples of the transformed PH tuples provided to the system in Stage 1.
\begin{table}[t]
    \centering
    \small
    \begin{tabular}{@{}p{0.7\linewidth}c@{}}
        \toprule
        \textbf{Premise (P), Hypothesis (H)} & \textbf{Label (L)}\\
        \midrule
        \textbf{P:} A man is being filmed in the middle of a soccer field. \newline
        \textbf{H:} A male is being recorded on a sports field. & Entailment\\ 
        \textbf{P:} A woman is talking on the phone while standing next to a dog. \newline
        \textbf{H:} The woman is sleeping in her room. & Contradiction\\
        \textbf{P:} A girl is talking on a cellphone. \newline
        \textbf{H:} The girl is calling her mother. & Neutral \\
        \bottomrule 
    \end{tabular}
    \caption{Illustrative examples of the NLI task for Entailment, Contradiction, and Neutral label.}
    \label{tab:nli_examples}
\end{table}
\begin{table*}[t]
\small
    \centering
    \resizebox{\linewidth}{!}{
    \begin{tabular}{@{}p{0.12\linewidth}>{\RaggedRight}p{0.44\linewidth}>{\RaggedRight}p{0.44\linewidth}@{}}
    \toprule
        \textbf{Label} &
        \textbf{Original Premise (P), Hypothesis (H)} &
        \textbf{Transformed Premise (P), Hypothesis (H)}
        \\
    \midrule
         
        Entailment & 
        \textbf{P}: A young family enjoys feeling ocean waves lap at their feet.  \newline
        \textbf{H}: A family is at the beach. & 
        
        \textbf{P}: A young family enjoys feeling ocean waves lap at their feet.  \newline
        \textbf{H}: A family is near the water.\\ 
        
        Entailment & 
        \textbf{P}: A woman with a green headscarf, blue shirt and a very big grin.  \newline
        \textbf{H}: The woman is very happy. & 
        
        \textbf{P}: A woman with a green headscarf, blue shirt and a very big grin.  \newline
        \textbf{H}: The woman is in a good mood.\\ 

        Neutral & 
        \textbf{P}: People jump over a mountain crevasse on a rope.  \newline
        \textbf{H}: Some people look visually afraid to jump. & 
        
        \textbf{P}: People are jumping over a crevasse.  \newline
        \textbf{H}: Some people look visually afraid to jump.\\ 
        
        Neutral & 
        \textbf{P}: A dog jumping for a Frisbee in the snow.  \newline
        \textbf{H}: A pet is enjoying a game of fetch with his owner. & 
        
        \textbf{P}: The dog is playing in the snow.  \newline
        \textbf{H}: A pet is enjoying a game of fetch with his owner.\\ 
        
        Contradiction & 
        \textbf{P}: Three firefighter come out of subway station.  \newline
        \textbf{H}: Three firefighters playing cards inside a fire station. & 
        
        \textbf{P}: Three firefighter come out of subway station.  \newline
        \textbf{H}: Three firefighters are inside a fire station.\\ 
        
        Contradiction & 
        \textbf{P}: An older women tending to a garden.  \newline
        \textbf{H}: The lady is cooking dinner. & 
        
        \textbf{P}: An older women tending to a garden.  \newline
        \textbf{H}:  The lady is making a meal. \\ 

    \bottomrule

    \end{tabular}
    }
    \caption{Illustrative examples corresponding to each label for Stage 1.}
    \label{tab:stage1_examples}
\end{table*}

\paragraph{Stage 2:}
In Stage 2, some knowledge statements relevant to the test instance are provided.
\begin{table*}[t]
\small
    \centering
    \resizebox{\linewidth}{!}{
    \begin{tabular}{@{}p{0.50\linewidth}>{\RaggedRight}p{0.42\linewidth}@{}}
    \toprule
        \textbf{Premise (P), Hypothesis (H)} &
        \textbf{Knowledge}
        \\
    \midrule
        \textbf{P:} Children bathe in water from large drums. \newline
        \textbf{H:} The kids are wet. &
         Water is related to wet \newline
         Water is a fluid \\ 
        \textbf{P:} Boys in what appears to be a library or school room. \newline
        \textbf{H:} Boys are in a place of learning. &
        You can use a library to obtain and read books. \newline
        A library is for finding information. \\
        
        \textbf{P:} Kids work at computers with a teacher's help. \newline
        \textbf{H:} The kids are learning. &
        A teacher wants students to learn. \newline
        Teacher is a type of educator. \\
        
        \textbf{P:} A couple walk hand in hand down a street. \newline
        \textbf{H:} A couple is sitting on a bench. &
        If you want to walk then you should stand \newline
        Walk is related to movement. \\

        \textbf{P:} A dog jumping for a Frisbee in the snow. \newline
        \textbf{H:} A cat washes his face and whiskers with his front paw. &
        Cat is not dog. \newline
        Paw is a type of animal foot. \\
    \bottomrule

    \end{tabular}
    }
    \caption{Examples showing top 2 knowledge statements provided in Stage 2.}
    \label{tab:stage2_examples}
\end{table*}
We collect these knowledge statements from ConceptNet~\cite{speer2017conceptnet} by querying for nouns and verbs present in the sentence and ranking based on the similarity. 
Table \ref{tab:stage2_examples} shows examples of knowledge statements fetched for SNLI test instances in Stage 2.

\paragraph{Stage 3:}
In Stage 3, we provide a few labeled examples similar to the test instance in case of abstention. We use POS tagger of spacy library \cite{spacy} and find examples with matching subjects and nouns. 
For each instance, we find similar examples from its corresponding training dataset.
Note that we sample equal number of examples for each label to avoid label imbalance.
We experiment varying the number of similar examples in this stage from $8$ to $128$.
\begin{table*}[t]
\small
    \centering
    \resizebox{\linewidth}{!}{
    \begin{tabular}{@{}p{0.12\linewidth}>{\RaggedRight}p{0.44\linewidth}>{\RaggedRight}p{0.44\linewidth}@{}}
    \toprule
        \textbf{Label} &
        \textbf{Premise} &
        \textbf{Hypothesis}
        \\
    \midrule
         
        Entailment & 
        A tattooed skateboarder is doing a trick.  &
        A tattooed skateboarder is pulling a stunt. \\ 
        
        Entailment & 
        The man is performing a trick on a bicycle high in the air.  &
        The man can ride a bike. \\ 
        
        Neutral & 
        A tattooed skateboarder is doing a trick.  &
        A skateboarder is performing for a crowd. \\ 
        
        Neutral & 
        A man on a bike tries to do a trick on the railing of an outdoor fountain.  &
        The man has on elbow pads and a helmet. \\ 
        
        Contradiction & 
        A man is pulling off a trick on his rollerblades. & 
        A man is sitting down on the floor.  \\
        
        Contradiction & 
        A boy doing a trick in the air on his bicycle. & 
        A boy is playing a clarinet.  \\
        

    \bottomrule

    \end{tabular}
    }
    \caption{Illustrative examples corresponding to each label for Stage 3 for the original test instance: \textbf{P}: A skateboarding youth does a trick on a rail., \textbf{H}: A young person on a skateboard.}
    \label{tab:stage3_examples}
\end{table*}
Table \ref{tab:stage3_examples} shows examples found for an SNLI test instance from the SNLI training dataset in Stage 3.

\paragraph{Stage 4:}
In Stage 4, we provide a number of unlabeled examples compiled from the corresponding training dataset of the test instance. 
We experiment varying the number of unlabeled examples in this stage from $5000$ to $20000$.
We expect this stage to be particularly beneficial for the OOD inputs as this provides exposure to instances that have not been observed during training performed prior to Stage 1.

\section{Experiments and Results}
In this section, we provide experimental details and analyse performance of our baseline approach.

\subsection{Approach}


\begin{figure}[t]
    \centering
    \includegraphics[width=\linewidth]{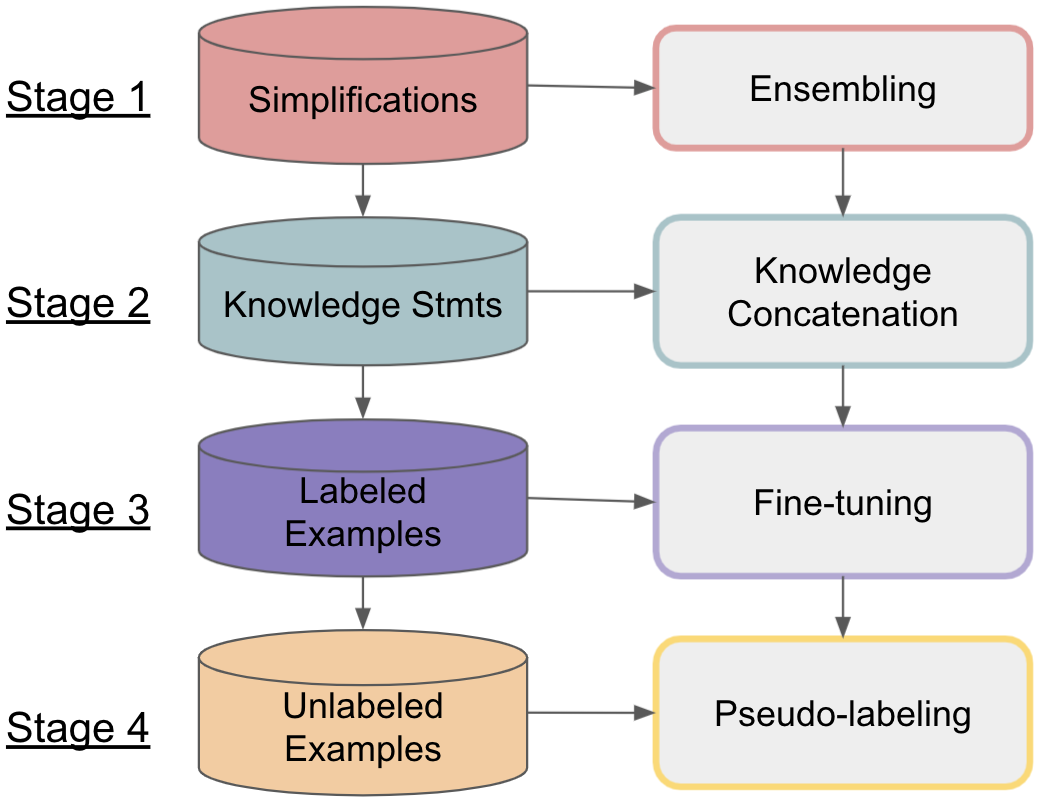}
    \caption{Our baseline approach for each stage of the proposed task.}
    \label{fig:approach}
    \end{figure}
We train a 3-way classification model on the training dataset and use $MaxProb$ i.e maximum softmax probability across the three classes as the confidence measure.
Initially, we make inference on the given test instance and proceed to Stage 1 if the confidence is below the threshold.
For the first stage, we further make inference on the provided simplifications of the test instance and find the most frequent prediction among those.
Then, we use \textbf{ensembling} techniques to find the final prediction i.e we compare the prediction on the original input with this most frequent prediction and if that prediction matches then we take the maximum prediction confidence among the variants where the system predicts the most frequent label otherwise we take the average of those prediction confidences.
For the second stage, we \textbf{concatenate the knowledge statement(s)} with the premise and make inference on the concatenated input.
For the third stage, we further \textbf{fine-tune} the model on the provided labeled instances and reattempt the original test instance using the fine-tuned model. 
For the final stage, we \textbf{pseudo-label} the provided unlabeled examples using the finetuned model obtained in Stage 3, fine-tune the model using those pseudo-labeled examples, and make inference on the test instance again. Figure \ref{fig:approach} illustrate our baseline approach for each stage of the proposed task.
\begin{table}[t]
    \centering
    \small 
    \begin{tabular}{p{1cm}p{0.9cm}p{0.9cm}p{0.9cm}p{0.9cm}}
    \toprule
        \textbf{Stage} & \textbf{SNLI}  & \textbf{MNLI mat.} &  \textbf{MNLI mis.}  & \textbf{DNLI}\\
        \midrule
        \textbf{S1} & 0.89	& -1.29	& 0.71	& 2.29 \\
        \textbf{S2} &    -7.62	& -2.01	& 0.37	& 1.91 \\
        \textbf{S3} &   -7.88	& 10.54	& 9.5	& 54.88 \\
        \textbf{S4*} &   -	& -	& -	& 72.02 \\
        
    \bottomrule
    \end{tabular}
    \caption{Table showing percentage improvement at various stages for all datasets. * indicates that this has been evaluated for 200 samples only due to limited computational budget.}
    \label{tab:auc_results}
\end{table}

\begin{figure*}[t]
\centering

    \begin{subfigure}{.46\textwidth}
        \includegraphics[width=\linewidth,height=46mm]{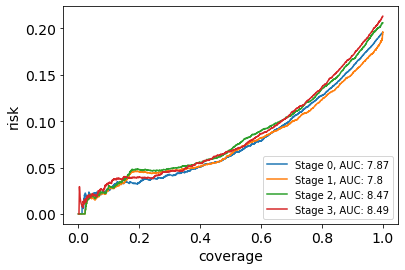}
        \caption{SNLI}
    \end{subfigure}
    \begin{subfigure}{.46\textwidth}
        \includegraphics[width=\linewidth,height=46mm]{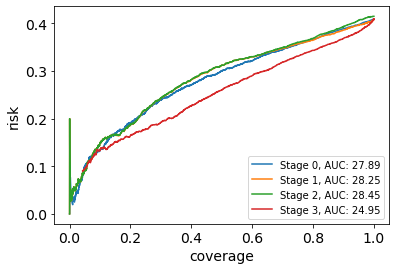}
        \caption{MNLI matched}
    \end{subfigure}
    \newline
    \begin{subfigure}{.46\textwidth}
        \includegraphics[width=\linewidth,height=46mm]{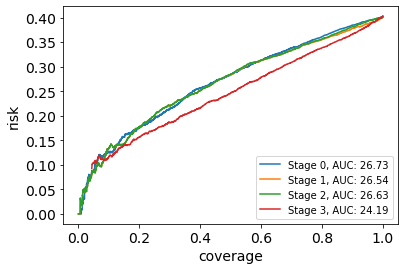}
        \caption{MNLI mismatched}
    \end{subfigure}
    \begin{subfigure}{.46\textwidth}
        \includegraphics[width=\linewidth,height=46mm]{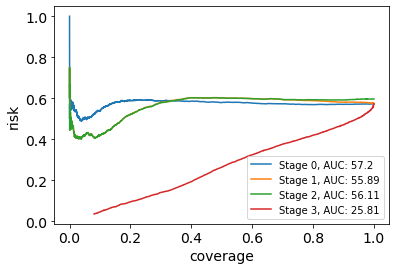}
        \caption{DNLI}
    \end{subfigure}
    \caption{Risk coverage curves for all datasets in Stages 0, 1, 2, and 3.}
    \label{fig:risk_coverage_curves}    
\end{figure*}
Table \ref{tab:stage0_results} shows the performance of the SNLI-trained model on all the evaluation datasets before Stage 1. We refer this stage as Stage 0 in our analysis.
As expected, the in-domain accuracy is high and AUC of risk-coverage curve is low. Whereas, the out-of-domain accuracy is low and AUC is high.
\subsection{Experimental Details}
Since NLI is a 3-way classification task, we use BERT-BASE model \cite{devlin-etal-2019-bert} with a linear layer on top of [CLS] token representation for training the model.  We use batch sizes of 32 and a learning rate ranging in $\{1{-}5\}e{-}5$. All experiments are done in Nvidia V100 16GB GPUs.
We train the model using 10k examples of the SNLI training dataset and evaluate on SNLI, MNLI (matched and mismatched), and DNLI datasets. 
\begin{table}[t]
    \centering
    \small 
    \begin{tabular}{p{1.7cm}p{0.9cm}p{0.9cm}p{0.9cm}p{0.9cm}}
    \toprule
        \textbf{Metric} & \textbf{SNLI}  & \textbf{MNLI mat.} &  \textbf{MNLI mis.}  & \textbf{DNLI}\\
        \midrule
        \textbf{Accuracy $\uparrow$} & 80.43 & 59.12 & 59.74 & 42.73\\
        \textbf{AUC $\downarrow$}  & 7.87 & 27.89 & 26.73 & 57.2\\
        
    \bottomrule
    \end{tabular}
    \caption{Table showing metric values obtained by the SNLI trained model on all the evaluation datasets before Stage 1 i.e without providing any extra information about the test instances. AUC corresponds to Area under risk-coverage curve. $\uparrow$ indicates higher is better while $\downarrow$ indicates lower is better.}
    \label{tab:stage0_results}
\end{table}

\subsection{MaxProb as a confidence Measure}
\begin{figure}[t]
    \centering
    \includegraphics[width=\linewidth]{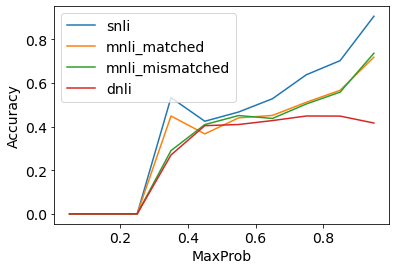}
    \caption{MaxProb Vs Accuracy plot for the SNLI trained model on all evaluation datasets.}
    \label{fig:conf_vs_acc}
\end{figure}

We plot $MaxProb$ vs $Accuracy$ achieved by the SNLI trained model for all the datasets in Figure \ref{fig:conf_vs_acc}.
It shows that $MaxProb$ is positively correlated with correctness i.e with increase in $MaxProb$, the accuracy also increases. 
This justifies the use of $MaxProb$ as a confidence measure for our task.

\subsection{Performance Prior to Stage 1}

\subsection{Performance Analysis}
Figure \ref{fig:risk_coverage_curves} shows the risk-coverage curves for all the evaluation datasets obtained in various stages of our task.
We find that Stage 3 leads to a significant improvement for OOD datasets.
In contrast, there is a marginal drop in performance for in-domain dataset (SNLI). 
This is expected as the model is already trained on the training dataset of the in-domain dataset and fine-tuning on a few examples in Stage 3 leads to overfitting and hence drop in performance. 
Furthermore, we find that their is not a significant improvement in performance in Stage 1 and Stage 2. This leaves scope for better ways to leverage the input simplifications and knowledge statements in Stage 1 and 2 respectively.

\section{Conclusion}
We introduced a multi-stage task in order to bridge the gap between real-world and standard NLP task formulations. 
Inspired by human-human interaction such as an interview,
we designed our task by incorporating various forms of guidance to help a system improve its prediction and learn the underlying concept to achieve generalization.
We instantiated the proposed task in Natural Language Inference setting and demonstrated that each of the stages improve OOD generalization performance of systems.
However, there still exists significant room to improve OOD generalization at each stage (especially Stage 1 and 2). 
We hope this work will bring more attention to developing NLP systems that align more closely with the real-world tasks.

\bibliography{tacl2018}

\begin{thebibliography}{47}
\expandafter\ifx\csname natexlab\endcsname\relax\def\natexlab#1{#1}\fi

\bibitem[{Alberti et~al.(2019)Alberti, Andor, Pitler, Devlin, and
  Collins}]{alberti-etal-2019-synthetic}
Chris Alberti, Daniel Andor, Emily Pitler, Jacob Devlin, and Michael Collins.
  2019.
\newblock \href {https://doi.org/10.18653/v1/P19-1620} {Synthetic {QA} corpora
  generation with roundtrip consistency}.
\newblock In \emph{Proceedings of the 57th Annual Meeting of the Association
  for Computational Linguistics}, pages 6168--6173, Florence, Italy.
  Association for Computational Linguistics.

\bibitem[{Anantha et~al.(2021)Anantha, Vakulenko, Tu, Longpre, Pulman, and
  Chappidi}]{anantha-etal-2021-open}
Raviteja Anantha, Svitlana Vakulenko, Zhucheng Tu, Shayne Longpre, Stephen
  Pulman, and Srinivas Chappidi. 2021.
\newblock \href {https://doi.org/10.18653/v1/2021.naacl-main.44} {Open-domain
  question answering goes conversational via question rewriting}.
\newblock In \emph{Proceedings of the 2021 Conference of the North American
  Chapter of the Association for Computational Linguistics: Human Language
  Technologies}, pages 520--534, Online. Association for Computational
  Linguistics.

\bibitem[{Bowman et~al.(2015)Bowman, Angeli, Potts, and
  Manning}]{bowman-etal-2015-large}
Samuel~R. Bowman, Gabor Angeli, Christopher Potts, and Christopher~D. Manning.
  2015.
\newblock \href {https://doi.org/10.18653/v1/D15-1075} {A large annotated
  corpus for learning natural language inference}.
\newblock In \emph{Proceedings of the 2015 Conference on Empirical Methods in
  Natural Language Processing}, pages 632--642, Lisbon, Portugal. Association
  for Computational Linguistics.

\bibitem[{Brown et~al.(2020)Brown, Mann, Ryder, Subbiah, Kaplan, Dhariwal,
  Neelakantan, Shyam, Sastry, Askell, Agarwal, Herbert-Voss, Krueger, Henighan,
  Child, Ramesh, Ziegler, Wu, Winter, Hesse, Chen, Sigler, Litwin, Gray, Chess,
  Clark, Berner, McCandlish, Radford, Sutskever, and
  Amodei}]{NEURIPS2020_1457c0d6}
Tom Brown, Benjamin Mann, Nick Ryder, Melanie Subbiah, Jared~D Kaplan, Prafulla
  Dhariwal, Arvind Neelakantan, Pranav Shyam, Girish Sastry, Amanda Askell,
  Sandhini Agarwal, Ariel Herbert-Voss, Gretchen Krueger, Tom Henighan, Rewon
  Child, Aditya Ramesh, Daniel Ziegler, Jeffrey Wu, Clemens Winter, Chris
  Hesse, Mark Chen, Eric Sigler, Mateusz Litwin, Scott Gray, Benjamin Chess,
  Jack Clark, Christopher Berner, Sam McCandlish, Alec Radford, Ilya Sutskever,
  and Dario Amodei. 2020.
\newblock \href
  {https://proceedings.neurips.cc/paper/2020/file/1457c0d6bfcb4967418bfb8ac142f64a-Paper.pdf}
  {Language models are few-shot learners}.
\newblock In \emph{Advances in Neural Information Processing Systems},
  volume~33, pages 1877--1901. Curran Associates, Inc.

\bibitem[{Chang et~al.(2020)Chang, Liu, Gopalakrishnan, Hedayatnia, Zhou, and
  Hakkani-Tur}]{chang-etal-2020-incorporating}
Ting-Yun Chang, Yang Liu, Karthik Gopalakrishnan, Behnam Hedayatnia, Pei Zhou,
  and Dilek Hakkani-Tur. 2020.
\newblock \href {https://doi.org/10.18653/v1/2020.deelio-1.9} {Incorporating
  commonsense knowledge graph in pretrained models for social commonsense
  tasks}.
\newblock In \emph{Proceedings of Deep Learning Inside Out (DeeLIO): The First
  Workshop on Knowledge Extraction and Integration for Deep Learning
  Architectures}, pages 74--79, Online. Association for Computational
  Linguistics.

\bibitem[{Chow(1957)}]{chow1957optimum}
Chi-Keung Chow. 1957.
\newblock An optimum character recognition system using decision functions.
\newblock \emph{IRE Transactions on Electronic Computers}, (4):247--254.

\bibitem[{Chung et~al.(2018)Chung, Lee, and Glass}]{chung-etal-2018-supervised}
Yu-An Chung, Hung-Yi Lee, and James Glass. 2018.
\newblock \href {https://doi.org/10.18653/v1/N18-1143} {Supervised and
  unsupervised transfer learning for question answering}.
\newblock In \emph{Proceedings of the 2018 Conference of the North {A}merican
  Chapter of the Association for Computational Linguistics: Human Language
  Technologies, Volume 1 (Long Papers)}, pages 1585--1594, New Orleans,
  Louisiana. Association for Computational Linguistics.

\bibitem[{Dasigi et~al.(2019)Dasigi, Liu, Marasovi{\'c}, Smith, and
  Gardner}]{dasigi-etal-2019-quoref}
Pradeep Dasigi, Nelson~F. Liu, Ana Marasovi{\'c}, Noah~A. Smith, and Matt
  Gardner. 2019.
\newblock \href {https://doi.org/10.18653/v1/D19-1606} {{Q}uoref: A reading
  comprehension dataset with questions requiring coreferential reasoning}.
\newblock In \emph{Proceedings of the 2019 Conference on Empirical Methods in
  Natural Language Processing and the 9th International Joint Conference on
  Natural Language Processing (EMNLP-IJCNLP)}, pages 5925--5932, Hong Kong,
  China. Association for Computational Linguistics.

\bibitem[{Desai and Durrett(2020)}]{desai-durrett-2020-calibration}
Shrey Desai and Greg Durrett. 2020.
\newblock \href {https://doi.org/10.18653/v1/2020.emnlp-main.21} {Calibration
  of pre-trained transformers}.
\newblock In \emph{Proceedings of the 2020 Conference on Empirical Methods in
  Natural Language Processing (EMNLP)}, pages 295--302, Online. Association for
  Computational Linguistics.

\bibitem[{Devlin et~al.(2019)Devlin, Chang, Lee, and
  Toutanova}]{devlin-etal-2019-bert}
Jacob Devlin, Ming-Wei Chang, Kenton Lee, and Kristina Toutanova. 2019.
\newblock \href {https://doi.org/10.18653/v1/N19-1423} {{BERT}: Pre-training of
  deep bidirectional transformers for language understanding}.
\newblock In \emph{Proceedings of the 2019 Conference of the North {A}merican
  Chapter of the Association for Computational Linguistics: Human Language
  Technologies, Volume 1 (Long and Short Papers)}, pages 4171--4186,
  Minneapolis, Minnesota. Association for Computational Linguistics.

\bibitem[{Dhingra et~al.(2018)Dhingra, Danish, and
  Rajagopal}]{dhingra-etal-2018-simple}
Bhuwan Dhingra, Danish Danish, and Dheeraj Rajagopal. 2018.
\newblock \href {https://doi.org/10.18653/v1/N18-2092} {Simple and effective
  semi-supervised question answering}.
\newblock In \emph{Proceedings of the 2018 Conference of the North {A}merican
  Chapter of the Association for Computational Linguistics: Human Language
  Technologies, Volume 2 (Short Papers)}, pages 582--587, New Orleans,
  Louisiana. Association for Computational Linguistics.

\bibitem[{El-Yaniv et~al.(2010)}]{el2010foundations}
Ran El-Yaniv et~al. 2010.
\newblock On the foundations of noise-free selective classification.
\newblock \emph{Journal of Machine Learning Research}, 11(5).

\bibitem[{Elazar et~al.(2021)Elazar, Kassner, Ravfogel, Ravichander, Hovy,
  Schutze, and Goldberg}]{Elazar2021MeasuringAI}
Yanai Elazar, Nora Kassner, Shauli Ravfogel, Abhilasha Ravichander, E.~Hovy,
  H.~Schutze, and Yoav Goldberg. 2021.
\newblock Measuring and improving consistency in pretrained language models.
\newblock \emph{ArXiv}, abs/2102.01017.

\bibitem[{Geifman and El-Yaniv(2017)}]{geifman2017selective}
Yonatan Geifman and Ran El-Yaniv. 2017.
\newblock Selective classification for deep neural networks.
\newblock \emph{arXiv preprint arXiv:1705.08500}.

\bibitem[{Hendrycks and Gimpel(2017)}]{hendrycks17baseline}
Dan Hendrycks and Kevin Gimpel. 2017.
\newblock A baseline for detecting misclassified and out-of-distribution
  examples in neural networks.
\newblock \emph{Proceedings of International Conference on Learning
  Representations}.

\bibitem[{Honnibal et~al.(2020)Honnibal, Montani, Van~Landeghem, and
  Boyd}]{spacy}
Matthew Honnibal, Ines Montani, Sofie Van~Landeghem, and Adriane Boyd. 2020.
\newblock \href {https://doi.org/10.5281/zenodo.1212303} {{spaCy:
  Industrial-strength Natural Language Processing in Python}}.

\bibitem[{Hwang et~al.(2020)Hwang, Bhagavatula, Bras, Da, Sakaguchi, Bosselut,
  and Choi}]{hwang2020comet}
Jena~D Hwang, Chandra Bhagavatula, Ronan~Le Bras, Jeff Da, Keisuke Sakaguchi,
  Antoine Bosselut, and Yejin Choi. 2020.
\newblock Comet-atomic 2020: On symbolic and neural commonsense knowledge
  graphs.
\newblock \emph{arXiv preprint arXiv:2010.05953}.

\bibitem[{Jiang et~al.(2020)Jiang, Xu, Araki, and
  Neubig}]{jiang-etal-2020-know}
Zhengbao Jiang, Frank~F. Xu, Jun Araki, and Graham Neubig. 2020.
\newblock \href {https://doi.org/10.1162/tacl_a_00324} {How can we know what
  language models know?}
\newblock \emph{Transactions of the Association for Computational Linguistics},
  8:423--438.

\bibitem[{Jones et~al.(2020)Jones, Sagawa, Koh, Kumar, and
  Liang}]{jones2020selective}
Erik Jones, Shiori Sagawa, Pang~Wei Koh, Ananya Kumar, and Percy Liang. 2020.
\newblock Selective classification can magnify disparities across groups.
\newblock \emph{arXiv preprint arXiv:2010.14134}.

\bibitem[{Kamath et~al.(2020)Kamath, Jia, and
  Liang}]{kamath-etal-2020-selective}
Amita Kamath, Robin Jia, and Percy Liang. 2020.
\newblock \href {https://doi.org/10.18653/v1/2020.acl-main.503} {Selective
  question answering under domain shift}.
\newblock In \emph{Proceedings of the 58th Annual Meeting of the Association
  for Computational Linguistics}, pages 5684--5696, Online. Association for
  Computational Linguistics.

\bibitem[{Lakshminarayanan et~al.(2017)Lakshminarayanan, Pritzel, and
  Blundell}]{NIPS2017_9ef2ed4b}
Balaji Lakshminarayanan, Alexander Pritzel, and Charles Blundell. 2017.
\newblock \href
  {https://proceedings.neurips.cc/paper/2017/file/9ef2ed4b7fd2c810847ffa5fa85bce38-Paper.pdf}
  {Simple and scalable predictive uncertainty estimation using deep ensembles}.
\newblock In \emph{Advances in Neural Information Processing Systems},
  volume~30. Curran Associates, Inc.

\bibitem[{Le~Scao and Rush(2021)}]{le-scao-rush-2021-many}
Teven Le~Scao and Alexander Rush. 2021.
\newblock \href {https://doi.org/10.18653/v1/2021.naacl-main.208} {How many
  data points is a prompt worth?}
\newblock In \emph{Proceedings of the 2021 Conference of the North American
  Chapter of the Association for Computational Linguistics: Human Language
  Technologies}, pages 2627--2636, Online. Association for Computational
  Linguistics.

\bibitem[{Lee et~al.(2017)Lee, Lee, Lee, and Shin}]{lee2017training}
Kimin Lee, Honglak Lee, Kibok Lee, and Jinwoo Shin. 2017.
\newblock Training confidence-calibrated classifiers for detecting
  out-of-distribution samples.
\newblock \emph{arXiv preprint arXiv:1711.09325}.

\bibitem[{Lewis et~al.(2019)Lewis, Denoyer, and
  Riedel}]{lewis-etal-2019-unsupervised}
Patrick Lewis, Ludovic Denoyer, and Sebastian Riedel. 2019.
\newblock \href {https://doi.org/10.18653/v1/P19-1484} {Unsupervised question
  answering by cloze translation}.
\newblock In \emph{Proceedings of the 57th Annual Meeting of the Association
  for Computational Linguistics}, pages 4896--4910, Florence, Italy.
  Association for Computational Linguistics.

\bibitem[{Mishra et~al.(2021)Mishra, Khashabi, Baral, and
  Hajishirzi}]{Mishra2021NaturalIB}
Swaroop Mishra, Daniel Khashabi, Chitta Baral, and Hanna Hajishirzi. 2021.
\newblock Natural instructions: Benchmarking generalization to new tasks from
  natural language instructions.
\newblock \emph{ArXiv}, abs/2104.08773.

\bibitem[{Mishra et~al.(2020)Mishra, Mitra, Varshney, Sachdeva, and
  Baral}]{Mishra2020TowardsQF}
Swaroop Mishra, A.~Mitra, Neeraj Varshney, Bhavdeep~Singh Sachdeva, and Chitta
  Baral. 2020.
\newblock Towards question format independent numerical reasoning: A set of
  prerequisite tasks.
\newblock \emph{ArXiv}, abs/2005.08516.

\bibitem[{Mishra and Sachdeva(2020)}]{Mishra2020DoWN}
Swaroop Mishra and Bhavdeep~Singh Sachdeva. 2020.
\newblock Do we need to create big datasets to learn a task?
\newblock In \emph{SUSTAINLP}.

\bibitem[{Mitra et~al.(2019)Mitra, Banerjee, Pal, Mishra, and
  Baral}]{Mitra2019ExploringWT}
A.~Mitra, Pratyay Banerjee, Kuntal~Kumar Pal, Swaroop Mishra, and Chitta Baral.
  2019.
\newblock Exploring ways to incorporate additional knowledge to improve natural
  language commonsense question answering.
\newblock \emph{ArXiv}, abs/1909.08855.

\bibitem[{Nachum et~al.(2018)Nachum, Gu, Lee, and Levine}]{nachum2018data}
Ofir Nachum, Shixiang~Shane Gu, Honglak Lee, and Sergey Levine. 2018.
\newblock Data-efficient hierarchical reinforcement learning.
\newblock \emph{Advances in Neural Information Processing Systems},
  31:3303--3313.

\bibitem[{Platt et~al.(1999)}]{platt1999probabilistic}
John Platt et~al. 1999.
\newblock Probabilistic outputs for support vector machines and comparisons to
  regularized likelihood methods.
\newblock \emph{Advances in large margin classifiers}, 10(3):61--74.

\bibitem[{Schick and
  Sch{\"u}tze(2021{\natexlab{a}})}]{schick-schutze-2021-exploiting}
Timo Schick and Hinrich Sch{\"u}tze. 2021{\natexlab{a}}.
\newblock \href {https://www.aclweb.org/anthology/2021.eacl-main.20}
  {Exploiting cloze-questions for few-shot text classification and natural
  language inference}.
\newblock In \emph{Proceedings of the 16th Conference of the European Chapter
  of the Association for Computational Linguistics: Main Volume}, pages
  255--269, Online. Association for Computational Linguistics.

\bibitem[{Schick and
  Sch{\"u}tze(2021{\natexlab{b}})}]{schick-schutze-2021-just}
Timo Schick and Hinrich Sch{\"u}tze. 2021{\natexlab{b}}.
\newblock \href {https://doi.org/10.18653/v1/2021.naacl-main.185} {It{'}s not
  just size that matters: Small language models are also few-shot learners}.
\newblock In \emph{Proceedings of the 2021 Conference of the North American
  Chapter of the Association for Computational Linguistics: Human Language
  Technologies}, pages 2339--2352, Online. Association for Computational
  Linguistics.

\bibitem[{Shen et~al.(2020)Shen, Mao, He, Long, Trischler, and
  Chen}]{shen-etal-2020-exploiting}
Tao Shen, Yi~Mao, Pengcheng He, Guodong Long, Adam Trischler, and Weizhu Chen.
  2020.
\newblock \href {https://doi.org/10.18653/v1/2020.emnlp-main.722} {Exploiting
  structured knowledge in text via graph-guided representation learning}.
\newblock In \emph{Proceedings of the 2020 Conference on Empirical Methods in
  Natural Language Processing (EMNLP)}, pages 8980--8994, Online. Association
  for Computational Linguistics.

\bibitem[{Shin et~al.(2020)Shin, Razeghi, Logan~IV, Wallace, and
  Singh}]{shin-etal-2020-autoprompt}
Taylor Shin, Yasaman Razeghi, Robert~L. Logan~IV, Eric Wallace, and Sameer
  Singh. 2020.
\newblock \href {https://doi.org/10.18653/v1/2020.emnlp-main.346}
  {{A}uto{P}rompt: {E}liciting {K}nowledge from {L}anguage {M}odels with
  {A}utomatically {G}enerated {P}rompts}.
\newblock In \emph{Proceedings of the 2020 Conference on Empirical Methods in
  Natural Language Processing (EMNLP)}, pages 4222--4235, Online. Association
  for Computational Linguistics.

\bibitem[{Speer et~al.(2017)Speer, Chin, and Havasi}]{speer2017conceptnet}
Robyn Speer, Joshua Chin, and Catherine Havasi. 2017.
\newblock Conceptnet 5.5: An open multilingual graph of general knowledge.
\newblock In \emph{Proceedings of the AAAI Conference on Artificial
  Intelligence}, volume~31.

\bibitem[{Sucholutsky and Schonlau(2021)}]{sucholutsky2021less}
Ilia Sucholutsky and Matthias Schonlau. 2021.
\newblock Less than one'-shot learning: Learning n classes from m< n samples.
\newblock In \emph{Proceedings of the AAAI Conference on Artificial
  Intelligence}, volume~35, pages 9739--9746.

\bibitem[{Talmor et~al.(2019)Talmor, Herzig, Lourie, and
  Berant}]{talmor-etal-2019-commonsenseqa}
Alon Talmor, Jonathan Herzig, Nicholas Lourie, and Jonathan Berant. 2019.
\newblock \href {https://doi.org/10.18653/v1/N19-1421} {{C}ommonsense{QA}: A
  question answering challenge targeting commonsense knowledge}.
\newblock In \emph{Proceedings of the 2019 Conference of the North {A}merican
  Chapter of the Association for Computational Linguistics: Human Language
  Technologies, Volume 1 (Long and Short Papers)}, pages 4149--4158,
  Minneapolis, Minnesota. Association for Computational Linguistics.

\bibitem[{Tam et~al.(2021)Tam, Menon, Bansal, Srivastava, and
  Raffel}]{Tam2021ImprovingAS}
Derek Tam, R.~R. Menon, M.~Bansal, Shashank Srivastava, and Colin Raffel. 2021.
\newblock Improving and simplifying pattern exploiting training.
\newblock \emph{ArXiv}, abs/2103.11955.

\bibitem[{Varshney et~al.(2020)Varshney, Mishra, and Baral}]{varshney2020s}
Neeraj Varshney, Swaroop Mishra, and Chitta Baral. 2020.
\newblock It's better to say" i can't answer" than answering incorrectly:
  Towards safety critical nlp systems.
\newblock \emph{arXiv preprint arXiv:2008.09371}.

\bibitem[{Wang and Jiang(2019)}]{wang-jiang-2019-explicit}
Chao Wang and Hui Jiang. 2019.
\newblock \href {https://doi.org/10.18653/v1/P19-1219} {Explicit utilization of
  general knowledge in machine reading comprehension}.
\newblock In \emph{Proceedings of the 57th Annual Meeting of the Association
  for Computational Linguistics}, pages 2263--2272, Florence, Italy.
  Association for Computational Linguistics.

\bibitem[{Wang et~al.(2018)Wang, Zhu, Torralba, and Efros}]{wang2018dataset}
Tongzhou Wang, Jun-Yan Zhu, Antonio Torralba, and Alexei~A Efros. 2018.
\newblock Dataset distillation.
\newblock \emph{arXiv preprint arXiv:1811.10959}.

\bibitem[{Welleck et~al.(2019)Welleck, Weston, Szlam, and
  Cho}]{welleck-etal-2019-dialogue}
Sean Welleck, Jason Weston, Arthur Szlam, and Kyunghyun Cho. 2019.
\newblock \href {https://doi.org/10.18653/v1/P19-1363} {Dialogue natural
  language inference}.
\newblock In \emph{Proceedings of the 57th Annual Meeting of the Association
  for Computational Linguistics}, pages 3731--3741, Florence, Italy.
  Association for Computational Linguistics.

\bibitem[{Williams et~al.(2018)Williams, Nangia, and
  Bowman}]{williams-etal-2018-broad}
Adina Williams, Nikita Nangia, and Samuel Bowman. 2018.
\newblock \href {https://doi.org/10.18653/v1/N18-1101} {A broad-coverage
  challenge corpus for sentence understanding through inference}.
\newblock In \emph{Proceedings of the 2018 Conference of the North {A}merican
  Chapter of the Association for Computational Linguistics: Human Language
  Technologies, Volume 1 (Long Papers)}, pages 1112--1122, New Orleans,
  Louisiana. Association for Computational Linguistics.

\bibitem[{Yang et~al.(2017)Yang, Hu, Salakhutdinov, and
  Cohen}]{yang-etal-2017-semi}
Zhilin Yang, Junjie Hu, Ruslan Salakhutdinov, and William Cohen. 2017.
\newblock \href {https://doi.org/10.18653/v1/P17-1096} {Semi-supervised {QA}
  with generative domain-adaptive nets}.
\newblock In \emph{Proceedings of the 55th Annual Meeting of the Association
  for Computational Linguistics (Volume 1: Long Papers)}, pages 1040--1050,
  Vancouver, Canada. Association for Computational Linguistics.

\bibitem[{Ye et~al.(2021)Ye, Lin, and Ren}]{Ye2021CrossFitAF}
Qinyuan Ye, Bill~Yuchen Lin, and Xiang Ren. 2021.
\newblock Crossfit: A few-shot learning challenge for cross-task generalization
  in nlp.
\newblock \emph{ArXiv}, abs/2104.08835.

\bibitem[{Zhang et~al.(2020)Zhang, Zhao, Saleh, and Liu}]{zhang2020pegasus}
Jingqing Zhang, Yao Zhao, Mohammad Saleh, and Peter Liu. 2020.
\newblock Pegasus: Pre-training with extracted gap-sentences for abstractive
  summarization.
\newblock In \emph{International Conference on Machine Learning}, pages
  11328--11339. PMLR.

\bibitem[{Zhang et~al.(2021)Zhang, Gong, and Choi}]{zhang2021knowing}
Shujian Zhang, Chengyue Gong, and Eunsol Choi. 2021.
\newblock Knowing more about questions can help: Improving calibration in
  question answering.
\newblock \emph{arXiv preprint arXiv:2106.01494}.

\end{thebibliography}
\bibliographystyle{acl_natbib}

\end{document}